\documentclass[conference]{IEEEtran}
\IEEEoverridecommandlockouts
\usepackage{cite}
\usepackage{amsmath,amssymb,amsfonts}
\usepackage{algorithmic}
\usepackage{graphicx}
\usepackage{textcomp}
\usepackage{xcolor}
\usepackage{siunitx}
\usepackage{makecell}

\usepackage{booktabs}
\def\BibTeX{{\rm B\kern-.05em{\sc i\kern-.025em b}\kern-.08em
    T\kern-.1667em\lower.7ex\hbox{E}\kern-.125emX}}
\begin{document}
\bstctlcite{IEEEexample:BSTcontrol}

\title{Automated Automotive Radar Calibration With Intelligent Vehicles\\
\thanks{Part of this work was financially supported by the Federal Ministry for Economic Affairs and Climate Action of Germany within the program "New Vehicle and System Technologies" (project LUKAS, grant number 19A20004F).}
}

\author{\IEEEauthorblockN{Alexander Tsaregorodtsev\IEEEauthorrefmark{1}, Michael Buchholz\IEEEauthorrefmark{1}, and Vasileios Belagiannis\IEEEauthorrefmark{2}}
\IEEEauthorblockA{\IEEEauthorrefmark{1}\textit{Institute for Measurement-, Control-, and Microtechnology} \\
\textit{Ulm University}, Germany\\
{\tt \{firstname\}}.{\tt \{lastname\}}@uni-ulm.de}
\IEEEauthorblockA{\IEEEauthorrefmark{2}\textit{Chair of Multimedia Communications and Signal Processing} \\
\centerline{\textit{Friedrich-Alexander-Universität Erlangen-Nürnberg}, Germany}\\
vasileios.belagiannis@fau.de}
}

\maketitle

\begin{abstract}
While automotive radar sensors are widely adopted and have been used for automatic cruise control and collision avoidance tasks, their application outside of vehicles is still limited. As they have the ability to resolve multiple targets in 3D space, radars can also be used for improving environment perception. This application, however, requires a precise calibration, which is usually a time-consuming and labor-intensive task. We, therefore, present an approach for automated and geo-referenced extrinsic calibration of automotive radar sensors that is based on a novel hypothesis filtering scheme. Our method does not require external modifications of a vehicle and instead uses the location data obtained from automated vehicles. This location data is then combined with filtered sensor data to create calibration hypotheses. Subsequent filtering and optimization recovers the correct calibration. Our evaluation on data from a real testing site shows that our method can correctly calibrate infrastructure sensors in an automated manner, thus enabling cooperative driving scenarios.
\end{abstract}

\begin{IEEEkeywords}
automatic calibration, roadside infrastructure, automotive radars
\end{IEEEkeywords}

\section{Introduction}
With the increasing availability of advanced driver-assistance system (ADAS) functionalities in commercial vehicles and the advancements in the field of automated driving, the research of assisting roadside infrastructure is slowly moving into focus. Roadside units (RSUs) already exist for tasks like traffic surveillance and monitoring, speed limit enforcement, toll collection, and construction site warning. The capability of such an infrastructure is actively being investigated for assisted automated driving and cooperative maneuver planning~\cite{buchholz2022int, mertens21}. To provide the information required for assisted driving, the RSUs need to be equipped with sensors capable of distinguishing different types of road users. Similar to automated vehicles, cameras, lidars, and radars can be used for such purposes, as all three sensor types are capable of detecting traffic participants. However, if such sensor equipment is to be deployed at real intersections, cost efficiency comes into play, as potentially hundreds of locations need to be equipped. This can become costly if complex and expensive sensors are used in the field. At the same time, sensor redundancy is desired, as the quality and robustness of the environment perception can be improved by fusing information from different sensors and sensor types. Therefore, it can be beneficial to use a combination of camera and radar sensors, as both are comparatively cheap compared to lidar sensors. Moreover, they are also already installed in commercial vehicles and thus are widely available. 

When building such a proposed camera-radar sensor infrastructure setup, it has to be precisely extrinsically calibrated before use in any assisted and cooperative driving scenarios. This is usually a manual and time-consuming process. Targets need to be manually placed in the infrastructure environment, then recordings have to be made, and often annotated to obtain a precise extrinsic calibration~\cite{kim2018rc}. Furthermore, there are currently no approaches for automatic and geo-referenced radar calibration. Additionally, existing approaches for camera-radar cross-calibration may not be applicable, if the sensor's field of view (FoV) is not overlapping with the FoV of other sensors or no cameras are present.
\begin{figure}
    \centering
    \includegraphics[width=0.475\textwidth]{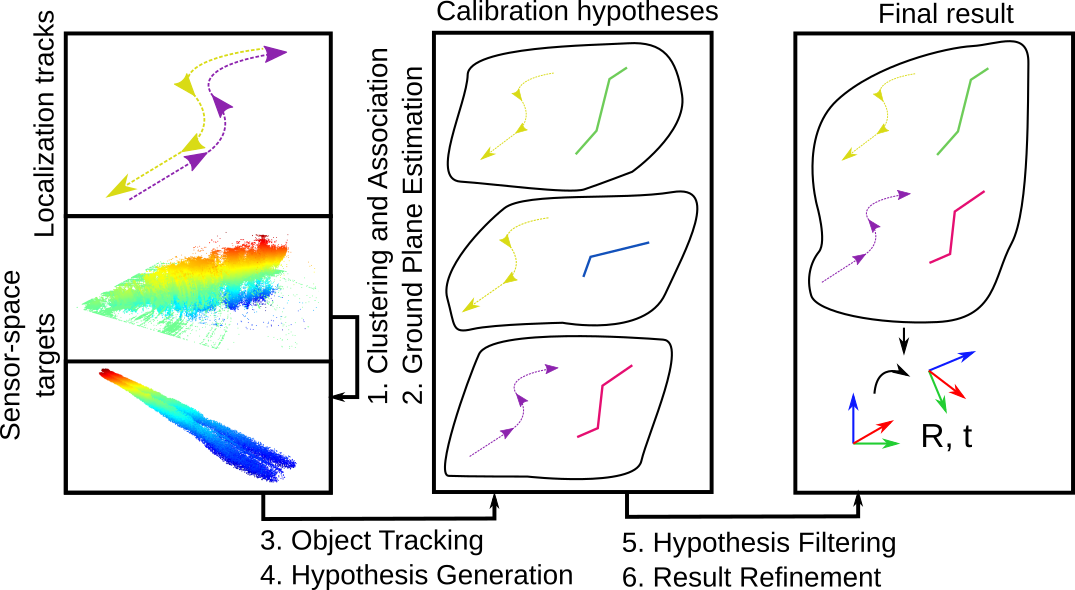}
    \caption{Overview of the calibration pipeline. Beginning on the left, the raw data is recorded and clustered (1). Afterward, the sensor roll and pitch are estimated by determining the ground plane (2). Using this information, object trajectories are smoothed and associated with vehicle localization data to create correspondences and estimate calibration hypotheses (3-4). The hypotheses are then filtered and refined to obtain the final calibration (5-6). Steps 1-6 are covered in detail in Sec.~\ref{sec:cluster}-\ref{sec:refinement}.}
    \label{fig:overview}
\end{figure}

We, therefore, propose a method for automatic infrastructure radar calibration, which uses cooperative and connected automated vehicles to provide a target vehicle with a known location, orientation, and dimension. An outline of our approach is shown in Fig.~\ref{fig:overview}. By recording radar target data in conjunction with connected vehicle data, a timestamped calibration dataset can be created. Then, using a multi-stage pipeline, the extrinsic calibration is estimated. In particular, objects are first detected in the raw sensor data and associated over time. From the obtained tracks, the driving lanes and the sensor pitch and roll are estimated, which allows the sensor data to be projected into a 2D plane. The object tracks are then re-tracked to improved accuracy and fused with the calibration vehicle data to obtain point-to-point correspondences, which are used to obtain possible sensor positions and rotations. By applying a novel filtering scheme and further refinement procedures, a final extrinsic calibration is obtained. Our contributions are summarized as:

\begin{itemize}
    \item An automated approach for geo-referenced infrastructure radar-only calibration based on the concept of hypothesis filtering relying on unmodified and unfiltered data from the infrastructure.
    \item The only requirements for the calibration vehicle are a localization unit and known vehicle dimensions. No appearance requirements for the vehicle as well as the sensor environment are imposed. Therefore other traffic participants may freely use the area observed by the sensor during the calibration process.
    \item Experiments on real-world data show that our method can be applied to radar sensors to be used to enhance the safety and reliability of automated and cooperative driving tasks.
\end{itemize}

\subsection{Prior Work}

The calibration of camera sensors is already widely explored for generic camera setups \cite{andrew2001multiple} as well as roadside infrastructure cameras \cite{datondji2016rotation} compared to radar sensors. Due to the different measurement type obtained from radars (3D target point clouds) compared to images (2D images with color and missing depth information), methods for estimating the 3D transformation~\cite{Briales_2017_CVPR} can be adapted to work with radars, while approaches solely relying on 2D-3D correspondences~\cite{lepetit2009epnp} may not be applicable.

Classical approaches for sensor calibration use human annotations and/or measurements of remarkable targets in both the real-world environment as well as the measurement space. Remarkable targets include chessboard patterns for camera, plastic spheres for lidar, and metal objects for radar sensors~\cite{kim2018rc}.
Feature-based approaches use sensor-specific features to perform matching with features from the real environment. Examples include the automatic calibration algorithm from \cite{dubska2014fully} based on line features and
\cite{ataer2014calibration} based on textured models.
~Other methods use features obtained from neural networks, e.g.~vehicle landmarks in
\cite{bartl2021automatic}, or semantic labels in 
\cite{ts2022semantic}.

More automatic calibration methods for outdoor applications are also available. The neural network presented in \cite{horn2020deepclr} performs a 3D point cloud registration between two point clouds, while~\cite{schoeller19camradar} uses neural networks to perform automatic rotational calibration between cameras and radars. 
\cite{li2023globally} optimizes a Gaussian mixture model to align radar target measurements to a GPS-equipped vehicle and estimate the yaw and 2D translation of infrastructure radars.

\section{Calibration Method}

We present our method as a multi-stage approach. Each stage is described in its own section and illustrated in Fig.~\ref{fig:overview}.

\subsection{Problem Formulation}

Consider the following transformation model:
\begin{equation}
    {\begin{bmatrix}
	x_\text{world} \\ y_\text{world} \\ z_\text{world}
	\end{bmatrix}} =
	\boldsymbol{R}
	{\begin{bmatrix}
	x_\text{sensor} \\ y_\text{sensor} \\ z_\text{sensor}
	\end{bmatrix}} + \boldsymbol{t}\;,
\end{equation}
where $[x_\text{sensor}, y_\text{sensor},z_\text{sensor}]^T$ are the coordinates of the radar sensor measurement in the sensor reference frame and $[x_\text{world}, y_\text{world}, z_\text{world}]^T$ is the measurement location in a global coordinate system. The goal of our approach is to estimate the sensor rotation matrix $\boldsymbol{R}$ and sensor position $\boldsymbol{t}$, where $\boldsymbol{R}$ can also be expressed by in terms of roll $\alpha$, pitch $\beta$, and yaw $\gamma$. The sensor-referenced measurements are radar targets $[x_\text{sensor}, y_\text{sensor},z_\text{sensor},v_\text{rad}]^T$, which are provided by the radar sensor. The world location in UTM
~coordinates together with the roll, pitch, and yaw of the calibration vehicle $[x_\text{world,UTM}, y_\text{world,UTM}, z_\text{world,UTM}, \alpha_\text{world}, \beta_\text{world}, \gamma_\text{world}]^T$ is obtained from a combined GNSS+RTK receiver and inertial measurement unit. Furthermore, measurements originating from both the radar sensor as well as the vehicle localization unit are assumed to be timestamped to the same time reference.

\subsection{Clustering and Association} \label{sec:cluster}

At the beginning, data from both the sensor and the vehicle's localization unit is recorded, while the vehicle passes through the sensor's field of view multiple times. This process is required to filter out false positives in the next steps. During the first data pre-processing step, the recorded radar targets are filtered in order to remove clutter detections. As the radar sensor itself is stationary, this is done by removing all targets which have a radial velocity of less than $0.1\si{km\per h}$ and thus are not considered to be moving. The remaining targets are then clustered based on an adapted DBSCAN algorithm~\cite{schubert2017dbscan}, which uses the euclidean distance between two data points to decide its association with an existing cluster.

By applying DBSCAN, multiple targets belonging to the same object can be grouped together. As this can cause problems with objects with low radar cross-section (RCS) values, e.g.~pedestrians, as well as objects, which pass by each other at a close distance, some adjustments have been made to the standard DBSCAN approach. To address the first problem, we set the minimum cluster size to 3 while accumulating the targets of three consecutive measurements. This allows us to identify moving objects which only produce one single target during a measurement step. To solve the second issue, we expanded the euclidean distance measure to also include the radial velocity $v_\text{rad}$ of each target. This increases the measured distance between two radar point clouds in close proximity if they are moving at different speeds and/or in different directions.

By using this modified approach, object clusters can be estimated for the entire calibration recording. However, the clusters between two consecutive time steps are still not associated with each other. Therefore, an association is implemented by exploiting the information gathered from the previously mentioned buffering procedure, where the cluster identifier of targets from previous timestamps is propagated to the current timestamp. Subsequently, the calculated associations are verified using the Hungarian Algorithm \cite{crouse2016hung}, where the distance between two cluster centers is used as the association cost. By introducing a gating step to prevent associating new object clusters with previous clusters that may already be outside the sensors' field of view, incorrect associations are removed. After this procedure, each clustered object in the target point cloud has an associated track it belongs to.

\subsection{Ground Plane Estimation}

In the next step, the pitch and roll of the sensor are estimated. As the sensor is installed on infrastructure, it may have a significant downward pitch in order to cover the desired road area. Therefore, the previously obtained object clusters can not be directly projected into the ground plane by removing the z-coordinate.
~Instead, we estimate the ground plane normal vector, which is then rotated to be the z-axis unit vector. The rotation required to do so contains both the pitch and the roll of the sensor. To estimate the 2D plane, we combine every target from the calibration sequence into one merged point cloud. Then, the point cloud density is estimated for each point by counting the number of neighboring points in a radius around the point. We discard all points which lie in low-density regions (in our case, we removed all points with less than 20\% of the maximum observed density) to obtain the most-occupied spaces which lie right above the lanes of the observed road area. Using the filtered point cloud, we can estimate its principal axes by performing a singular value decomposition (SVD)~\cite{golub1971singular} of the entire point cloud and using the left-hand base vector belonging to the smallest eigenvalue as the plane normal vector $\boldsymbol{n}$. By defining the desired normal vector to be $\boldsymbol{n}_\text{ref}=[0, 0, 1]^T$, we can calculate a rotation matrix $\boldsymbol{R}_{\alpha\beta}$ such that $\boldsymbol{n}_\text{ref}=\boldsymbol{R}_{\alpha\beta}\boldsymbol{n}$, which contains the roll and pitch of the sensor. The yaw is still unknown, as the ground plane only defines two orthogonal base vectors, against which the pitch and roll are estimated.

\subsection{Object Tracking}

After determining the roll and pitch of the sensor, the recorded targets can be transformed into a new sensor coordinate system with zero roll and pitch. By dismissing the z-coordinate 
we can transform the sensor data into a 2D bird's eye view (BEV) coordinate system. This allows us to track the objects using planar kinematic models. The tracking step is required, as the extent of the target cluster as well as the number of targets contained within a cluster changes with every measurement. This leads to extremely noisy object trajectories if the cluster center is used as a reference point. We use the Unscented Kalman filter~\cite{wan2000unscented} with a CTRA transition model~\cite{cvmodel} to smooth the cluster center trajectories. As we also have data from future time steps available for each measurement, we employ the Rauch-Tung-Striebel (RTS) Kalman smoother \cite{sarkka2008unscented} to further smooth the resulting trajectory by applying backward filtering. At the end of the tracking procedure, each object cluster in the recorded sequence is smoothed down to a trajectory containing the x- and y-positions of the object centers. 

\subsection{Hypothesis Generation and Filtering}

To choose the correct object track, we create calibration hypotheses to then confirm or reject them and find the final calibration. A calibration hypothesis is defined by a tuple containing an object track as well as the localization data from the vehicle. For each state from the object track, its corresponding timestamp is obtained from the original radar measurement. Afterward, the localization measurement with the closest timestamp is taken as the 3D correspondence, where the elevation of the object data can either be set to 0 or obtained from the targets prior to Kalman smoothing. Therefore, it is important that the time synchronicity of both data sources is ensured. Using these correspondences, a 3D transform can be estimated, e.g.~by employing convex global registration~\cite{Briales_2017_CVPR}. By repeating this step for each separate hypothesis, we obtain a transform for each one. These hypotheses can then be grouped by their estimated translation vector $\boldsymbol{t}$ using DBSCAN. As the calibration vehicle passed the sensor multiple times and thus is responsible for multiple separate object tracks, a cluster of correct sensor positions should exist. However, other location clusters may exist. To filter out incorrect location clusters, the angle of the rotation error matrix $\boldsymbol{R}_j^T\boldsymbol{R}_i$ converted to an axis-angle representation is calculated for every possible pair of transforms inside of the hypothesis cluster, as ideally $\boldsymbol{R}_i$ and $\boldsymbol{R}_j$ are the same, thus $\boldsymbol{R}_j^T\boldsymbol{R}_i=\boldsymbol{I}$. The cluster with the lowest average angle can then be chosen as the accepted calibration hypothesis by re-estimating the transform using correspondences merged from each track in the cluster, as the relative rotations of each hypothesis pair in the chosen cluster are closest to 0. The final calibration is then found by concatenating this estimated transform with the previously estimated $\boldsymbol{R}_{\alpha\beta}$.

\subsection{Result Refinement} \label{sec:refinement}

To further refine the result, the object tracking step can be repeated with the final transformation already applied to the target clusters. However, as the sensor location is already known, instead of using the cluster center, the target closest to the sensor can be chosen instead. By then incorporating the offset from the localization unit to the car point closest to the sensor, the dimensions of the vehicle are taken into account, and the noise caused by varying cluster sizes and extents is partially suppressed. This improves the quality of the correspondences and provides a better result. In the final step, the overall transform can be further refined by maximizing the number of targets contained in the ground area polygon $\mathbf{a}$ covered by the vehicle. For this purpose, we define a loss $\mathcal{L}(p)=\{\min_i|\mathbf{a}_i - p|\,\text{if}\,p\in\mathbf{a}, 0\, \text{else}\}$, where $\mathbf{a}_i$ is the $i$-th polygon vertex. Then, by accumulating the loss for all correspondences and using a gradient-free optimization method like~\cite{nelder1965simplex} to minimize the loss, we estimate a 2D offset vector, which, when applied after the calibration transform, corrects the calibration in such a way that most targets lie on the calibration vehicle.

\section{Evaluation}

We present an experimental evaluation based on recorded sensor data from our real-world test site.

\subsection{Data Sources and Evaluation Procedure}

To gather data for evaluation, we use the sensor setup of our real-world test site located in Ulm-Lehr \cite{buchholz2022int}. The test site contains a total of 7 light poles equipped with camera and automotive radar sensors and attached processing units. The radars mounted at the intersection consist of multiple long-range radars capable of outputting target lists with points in spherical coordinates in conjunction with the radial velocity and additional information about each point with a frequency of 15\si{Hz}. The calibration vehicle is equipped with a highly precise localization unit containing both a 4-band GNSS receiver and RTK correction support as well as an inertial measurement unit, whose position is calibrated with respect to the vehicles' rear axle position and the vehicle dimensions to model an occupancy 3D bounding box of the vehicle. The localization data is recorded with a frequency of 50 Hz and reaches a positional accuracy of up to 2\si{cm}. Both the radar and localization data are timestamped with the sensor clocks being synchronized to GPS time, with measured clock offsets below 2\si{ms}. For the evaluation protocol, two sequences of around 15 minutes in length are recorded for each sensor in parallel while the calibration vehicle passes all sensors' fields of view multiple times.

We run our calibration approach with the recorded data, where one sequence is used for calibration and the second one for evaluation. Only the object tracking step is performed on the evaluation set. Employing the previously used data association by synchronized timestamps, each object track is associated with a localization track and transformed into world coordinates by applying the previously estimated transform.
To remove tracks of other vehicles, associations with mean offsets between localization and radar data of over 5\si{m} are removed. The remaining tracks are merged into a single track. Due to our object clustering approach not estimating any object bounding boxes, we report the mean outlier error $\delta_{op}$, which is calculated by averaging the distance of all outlier targets to the vehicle polygons.
~We also introduce the metric $r_i$, which calculates the fraction of radar targets in a cluster lying inside the calibration vehicle's bounding box. $r_i$ captures the quality of the estimated alignment and considers the random nature of the occurring target point clouds. We furthermore choose the object cluster target closest to the camera and report the observed offset to the vehicle corner closest to the estimated sensor position $\delta_{p}$. However, it contains a random offset due to the randomness of the target location on the vehicle.

\subsection{Evaluation results}

The quantitative evaluation results for 4 sensors from our test site are presented in Table~\ref{tbl:real}.

\begin{table}[h]
    \caption{Quantitative evaluation results from real-world sequences.}
    \label{tbl:real}
    \centering
    \begin{tabular}{lcccc}
        \toprule
        \textbf{Radar name} & West & Center & East & West 2\\
        \midrule
        $r_i$ (\%) & 67.69 & 69.92 & 79.06 & 60.36\\
        \midrule
        $\delta_{op}$ (\si{m}) & 0.35 & 0.26 & 0.29 & 0.39\\
        \midrule
        $\delta_{p}$ (\si{m}) & 1.58 & 1.34 & 1.10 & 1.71\\
        \bottomrule
    \end{tabular}
\end{table}

\begin{figure}
    \centering
    \includegraphics[width=0.35\textwidth]{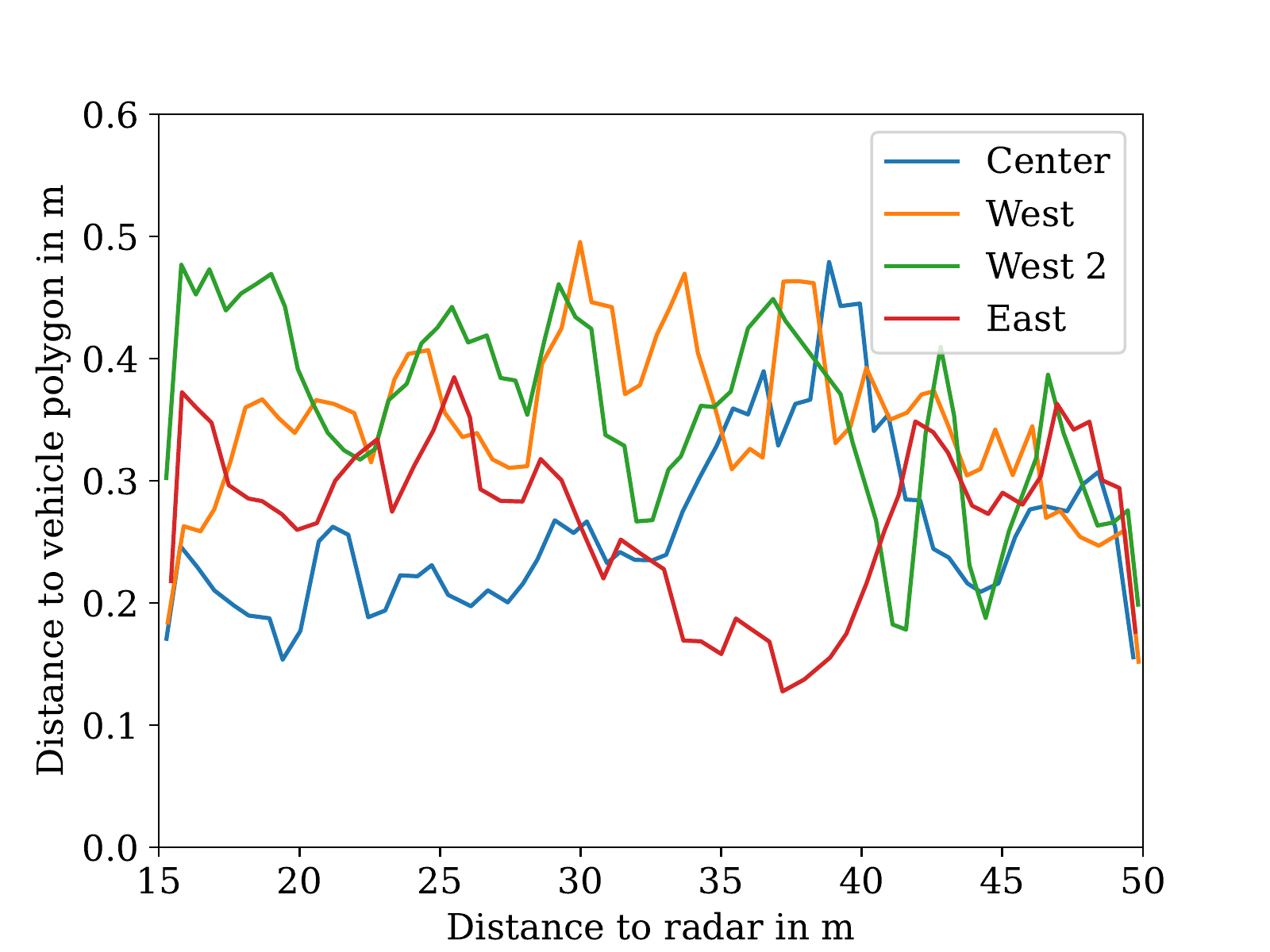}
    \caption{Distance $\delta_{op}$ between vehicle polygon and targets not contained in the vehicle polygon, which are considered outlier targets, plotted over the distance between target and sensor.}
    \label{fig:outlier}
\end{figure}

\begin{figure}
    \centering
    \includegraphics[width=0.35\textwidth]{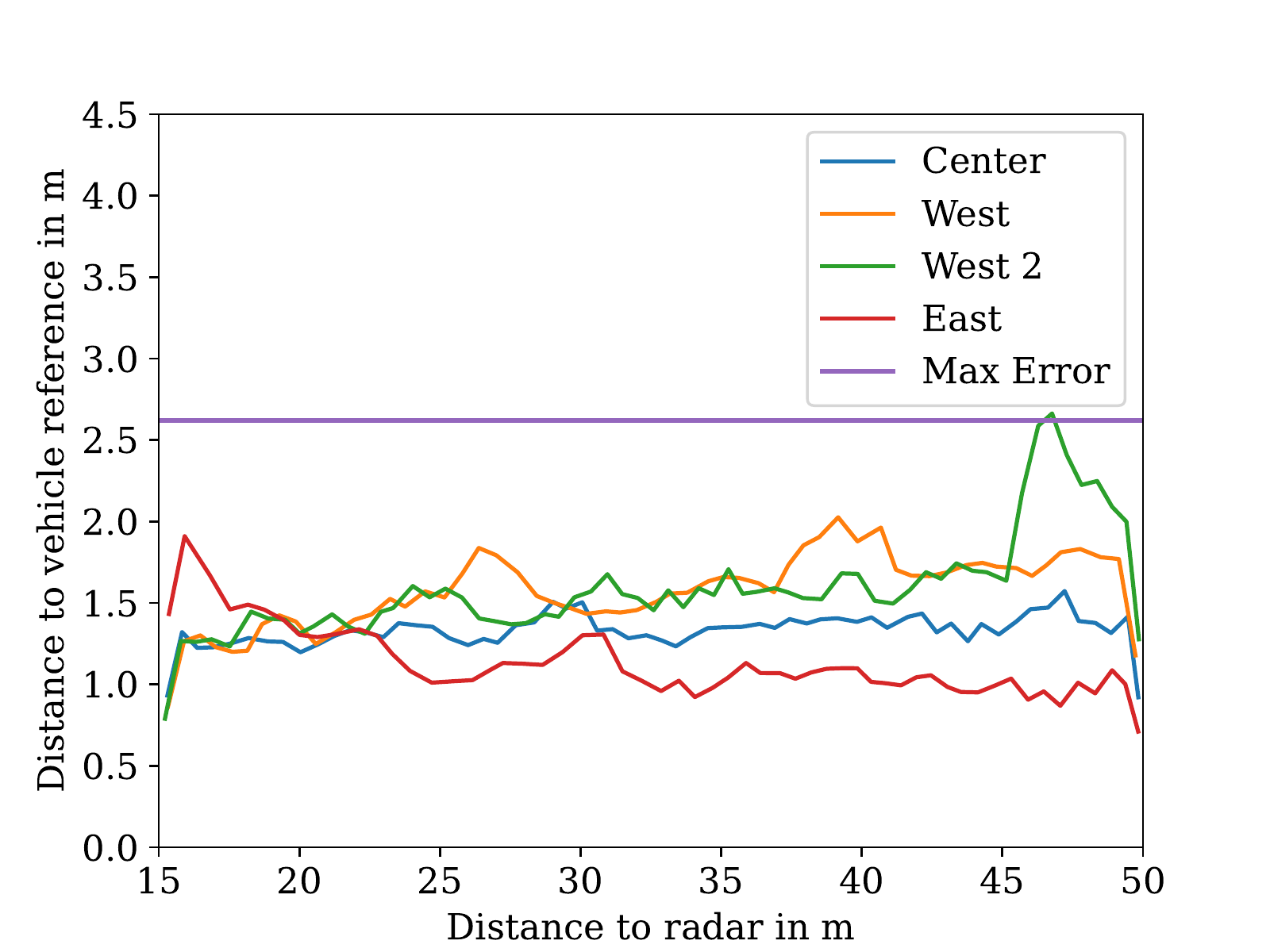}
    \caption{Distance $\delta_{p}$ between vehicle corner and target closest to the sensor plotted over the distance between sensor and vehicle corner. As a comparison, the distance from the vehicle corner to the vehicle geometric center (``Max error'') is visualized alongside the data to provide an upper bound for the acceptable error.}
    \label{fig:errors}
\end{figure}

\begin{figure}
    \centering
    \includegraphics[width=0.40\textwidth]{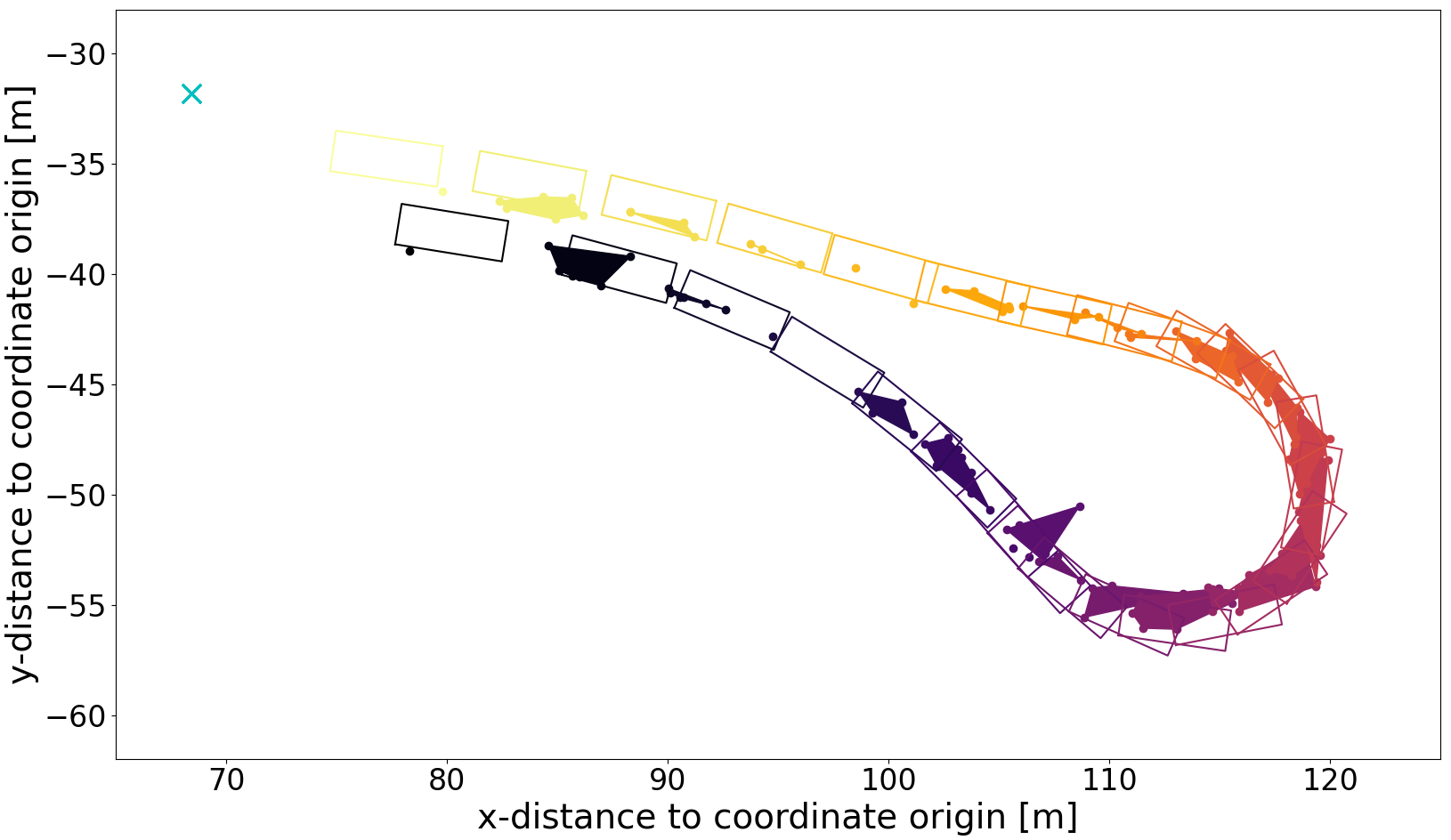}
    \caption{Bird's eye view of a sample test track. The boxes denote the calibration vehicle, while the points represent the associated target clusters. Clusters with at least three points are visualized using their convex hull for better clarity. Earlier measurements are colored black, while recent measurements are shown in yellow. The estimated sensor position is the cyan cross in the top left corner of the image.}
    \label{fig:bev}
\end{figure}

The results show the effectiveness of our approach despite evaluating on unstructured radar data. When considering both the inlier ratio $r_i$ and outlier distance $\delta_{op}$, we can observe ratios between 60\% and 70\%, while the remaining 30-40\% of targets have an average distance of below 50\si{cm} to the vehicle bounding box across all evaluated sensors with different fields of view, as shown in Fig.~\ref{fig:outlier}. As the raw radar data may still contain clutter targets due to unwanted reflections and measurement noise, parts of the outliers may be caused by clutter measurements and not caused by calibration errors. Additionally, average errors below 50\si{cm} provide enough precision for cooperative driving tasks.

While the average errors $\delta_{p}$ are higher compared to the outlier error $\delta_{op}$, they are mostly caused by the missing vehicle reference of the target cluster data, as a single target can originate from any reflective part of the vehicle. This is also apparent in Fig.~\ref{fig:errors}, where the distance between the chosen reference target and vehicle corner is visualized. As the error is similar between all evaluated sensors and nearly constant over the entire sensor range, systematic errors in the estimated sensor rotation are unlikely, as an incorrectly determined rotation causes offsets proportional to the sensor distance. 
Fig.~\ref{fig:bev} provides a sample visual result of the estimated calibration. As the sensor is located in the top left corner of the image, it mostly captures the left or back side of the vehicle, resulting in an increased probability of targets originating from the left or back side of the vehicle. Additionally, we can see that target clusters overlap well with the bounding box in most situations, excluding clutter targets.

\section{Conclusion}

We presented an approach for automated calibration of automotive radar sensors that is based on hypothesis filtering. Our method makes minimal assumptions about the calibration vehicle and the sensor environment, while it does not depend on human interaction. The calibration is performed by recording data from both the radar and the target vehicle by driving through the sensor's FoV and extracting object tracks from the recording. The sensor and vehicle tracks are combined during hypothesis generation. Subsequent filtering recovers the correct hypothesis, which is then further optimized to take the random nature of target locations into account. Our evaluation on real-world data shows the applicability of our approach. 
%


\bibliographystyle{ieeetran}
\bibliography{bibliography}

\end{document}